\documentclass[journal]{IEEEtran}
\usepackage{cite}
\usepackage{times,amsmath,amssymb,psfrag,stfloats}
\usepackage{tabularx}
\usepackage{graphicx}
\usepackage{epsfig}
\usepackage{epstopdf}
\graphicspath{{./pics/}} \DeclareGraphicsExtensions{.pdf,.jpg,.png,.eps}
\usepackage{color}
\usepackage{algorithmic}
\usepackage{algorithm}
\usepackage{hyperref}
\usepackage{cleveref}
\usepackage{multirow}
\usepackage{caption}
\usepackage{flushend}
\usepackage{algorithmic}
\usepackage{algorithm}
\renewcommand{\algorithmicrequire}{\textbf{Initialization:}}
\renewcommand{\algorithmicensure}{\textbf{Iteration:}}

\renewcommand{\baselinestretch}{0.95}

\begin{document}
\title{An Intelligent Question Answering System based on Power Knowledge Graph}

\author{\IEEEauthorblockN{Yachen Tang\IEEEauthorrefmark{1}, Haiyun Han, Xianmao Yu\IEEEauthorrefmark{2}, Jing Zhao\IEEEauthorrefmark{2}, Guangyi Liu\IEEEauthorrefmark{1}, and Longfei Wei\IEEEauthorrefmark{3}}\\
\IEEEauthorblockA{\IEEEauthorrefmark{1}Envision Digital, Redwood City, CA, USA},\\
\IEEEauthorblockA{\IEEEauthorrefmark{2}State Grid Sichuan Electric Power Company, Chengdu, Sichuan, China},\\
\IEEEauthorblockA{\IEEEauthorrefmark{3}Hitachi ABB Power Grids, San Jose, CA, USA}\\}

\maketitle

\begin{abstract}
The intelligent question answering (IQA) system can accurately capture users' search intention by understanding the natural language questions, searching relevant content efficiently from a massive knowledge-base, and returning the answer directly to the user. Since the IQA system can save inestimable time and workforce in data search and reasoning, it has received more and more attention in data science and artificial intelligence. This article introduced a domain knowledge graph using the graph database and graph computing technologies from massive heterogeneous data in electric power. It then proposed an IQA system based on the electrical power knowledge graph to extract the intent and constraints of natural interrogation based on the natural language processing (NLP) method, to construct graph data query statements via knowledge reasoning, and to complete the accurate knowledge search and analysis to provide users with an intuitive visualization. This method thoroughly combined knowledge graph and graph computing characteristics, realized high-speed multi-hop knowledge correlation reasoning analysis in tremendous knowledge. The proposed work can also provide a basis for the context-aware intelligent question and answer.
\end{abstract}

\IEEEpeerreviewmaketitle

\section{Introduction}
\IEEEPARstart{T}{HE} explosive growth of information and the continuous accumulation of redundant information has dramatically increased the difficulty of obtaining valuable information accurately and quickly. The knowledge graph describes the complex relationships between concepts and entities in the objective world in a structured form and expresses complex related information closer to human cognition. The knowledge graph usually uses graph databases to store knowledge; thus, the knowledge query process can be regarded as graph traversal and clustering. Based on the knowledge graph, an intelligent question answering (IQA) system combines the intent and understanding of the user's question with knowledge search, which can directly answer without manual intervention, saving much time and human resources for data search, which is getting more and more attention.

The open-domain question answering system deals with a wide range of problems, but the question structure and types are single, and the feedback answers are relatively simple. The user target of the domain knowledge graph is professionals of various levels in a specific industry. The accuracy requirements of knowledge search are high. The search results are usually used to assist in multiple complex analysis applications or decision support. Questions in a question answering system are generally complex, including statistics, reasoning, planning, and other functions. Therefore, the content and expression of feedbacks should be more diverse. The IQA system's design usually requires the domain knowledge graph's professional knowledge structure and in-depth analysis of user questions to achieve efficient and accurate questions and reasoning results.

General IQA technology has been relatively mature, and there are many commercial products, such as Apple's Siri. Researchers have conducted some research reviews on retrieving and evaluating the general question answering systems \cite{1, 2}. The authors in \cite{3} surveyed the task scope of open-domain text questions and answers and then summarized the latest critical developments of the open-domain IQA system based on deep learning. Aiming at the core technology in the IQA system, Chang and Levy elaborated on Chinese NLP's crucial technology based on Chinese syntax analysis and semantic understanding \cite{4, 5}. According to the knowledge base of the dynamic memory network, authors in \cite{6} realized the IQA system's answer. To improve the reasoning accuracy of question and answer, Lin et al. proposed SmartQ: a strategy for recommending suggested solutions from similar questions to each new question \cite{7}, and Zhao et al. used the social network via graph regularized matrix to infer the user model to improve the performance of expert finding in IQA system \cite{8}. Aiming at the IQA system based on the knowledge graph, authors in \cite{9} according to the constructed manufacturing knowledge graph, applied knowledge reuse to realize the IQA of product knowledge. Also, based on large-scale knowledge graphs, Luo et al. used closed reasoning of incremental theory to provide a theoretical basis for IQA \cite{10}. Besides, Hu and Md. Saiful Islam realized answering natural language questions through subgraph and similarity matching methods based on the knowledge graph \cite{11, 12}.

The transformation of the IQA system from generality to professionalism requires further excavation of industry background and professional knowledge and techniques in specific scenarios. The IQA system is close to practical in specific limited fields, for example, IBM's Watson in the medical and financial areas, Amazon's Alexa in the smart home and retail fields, and Microsoft's Cortana in the operating system assistant. However, the current IQA system for the electric power field is still in its infancy, and the relevant references are limited to the research on intelligent customer service \cite{13}, fault diagnosis \cite{14}, and the equipment knowledge graph in electric power systems \cite{15}. There are massive data collected during the entire electrical business management process, which includes planning, construction, commissioning, operation, maintenance, overhaul, and scrapping \cite{19,20}. According to the collected data, the authors will construct a knowledge graph of the power field first to integrate multiple data sources as one inter-connected graph. Then the development of an IQA system based on the domain knowledge graph can 1) help business users obtain the critical information they need from the massive power data efficiently and quickly; 2) realize the association analysis and associative reasoning during the data query to provide multiple rounds of question and answer.

The main contribution of this paper is to introduce an IQA system based on the power knowledge graph, which proposed 
\begin{enumerate}
    \item a method of in-depth analysis of questions based on semantic structure;
    \item a design of a power knowledge graph based on the ontology model (schema);
    \item an outline of a knowledge reasoning algorithm based on the schema;
    \item a complete set of an IQA system based on the power knowledge graph;
    \item the testing and application verification based on the real environment of power companies and related business departments.
\end{enumerate}

\noindent The rest of this paper is structured as follows: Section II introduced the framework of the IQA system and the ontology model of the power knowledge graph. Section III introduces the construction of the IQA system and Section IV represents the case studies. Section V concludes with future work.

\section{Research Foundation}
\subsection{Framework of IQA system}
\begin{figure}[h!bt]
\centerline{\includegraphics[scale = 0.45] {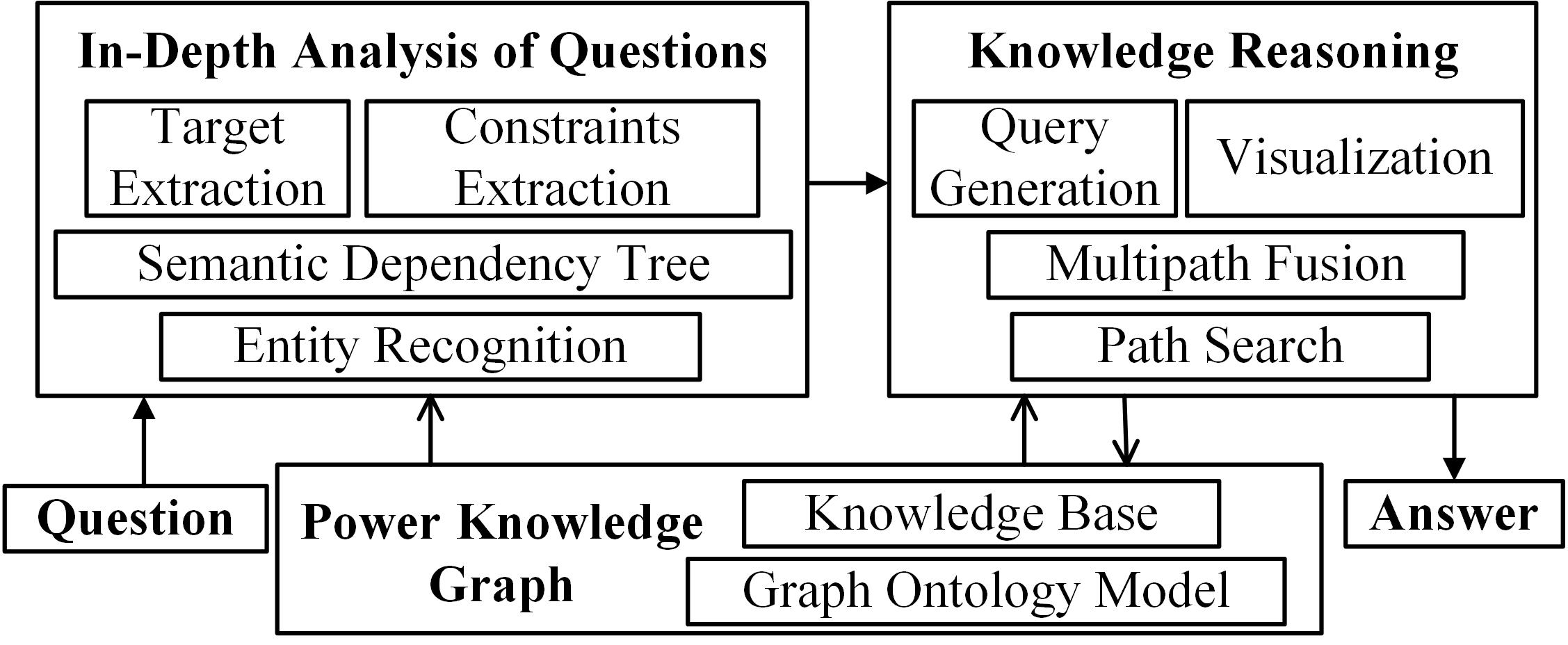}}
\caption{Framework of IQA system.}\label{figure1}
\end{figure}

The IQA system proposed in this paper is oriented to query knowledge in the electric power field, which used the rich information provided by the knowledge graph ontology model and combined the information with NLP technology. The framework of the system is illustrated in Fig. \ref{figure1}. 

The four main steps in the in-depth analysis of questions are performed simultaneously. Entity recognition can provide a comprehensive vocabulary to achieve accurate words segmentation (target and constrain extraction) based on the dependency tree. The extraction results will be recognized as entities to the constructed ontology model. As mentioned, the IQA system is designed based on a constructed power knowledge graph. First, it performs the in-depth analysis of the question sentence, which implements the graph ontology model and specific knowledge to identify entities in the question sentence, and then carries out semantic analysis to extract the target and constraint conditions. Next, the system executes the knowledge reasoning: use the graph ontology model to search and merge the reasoning path, automatically generate a query sentence, and run the query in the knowledge base. Finally, the query result can be visualized and the answer is returned to the user.

From this framework, the key to the in-depth analysis process of question sentences is the accuracy of target and constraint extraction, which directly determines whether the system can infer the correct path. The basis of knowledge reasoning is to improve the efficiency of path search and query execution so as to demonstrate user answers in real-time.

\subsection{Ontology Model of Power Knowledge Graph}
From the enterprise-level perspective, the power knowledge graph's ontology model is the unified modeling of the original data from each electric power company's business department. It is the framework and basis for constructing the electric power knowledge graph. The ontology model's rationality and accuracy will directly affect the accuracy and speed of knowledge queries in the IQA system. The ontology model proposed in this paper mainly contains two parts: the abstract model (for management purpose) and the physical model (to define the specific device). The design scope is not composed all kinds of statistical and index data. The abstract model is a unified definition of the power company's core business departments and their corresponding attributes and interrelationships. This design mainly avoids the repeated explanation of the same business object during the construction of the information system model from different departments, which will cause repeated data entry and maintenance. The physical model defines specific power equipment, clarifies the topological connections of various equipment, and describes subordination relationships between equipment and abstract models such as manufacturers, suppliers, management departments, and substations.

\begin{figure}[h!bt]
\centerline{\includegraphics[scale = 0.32] {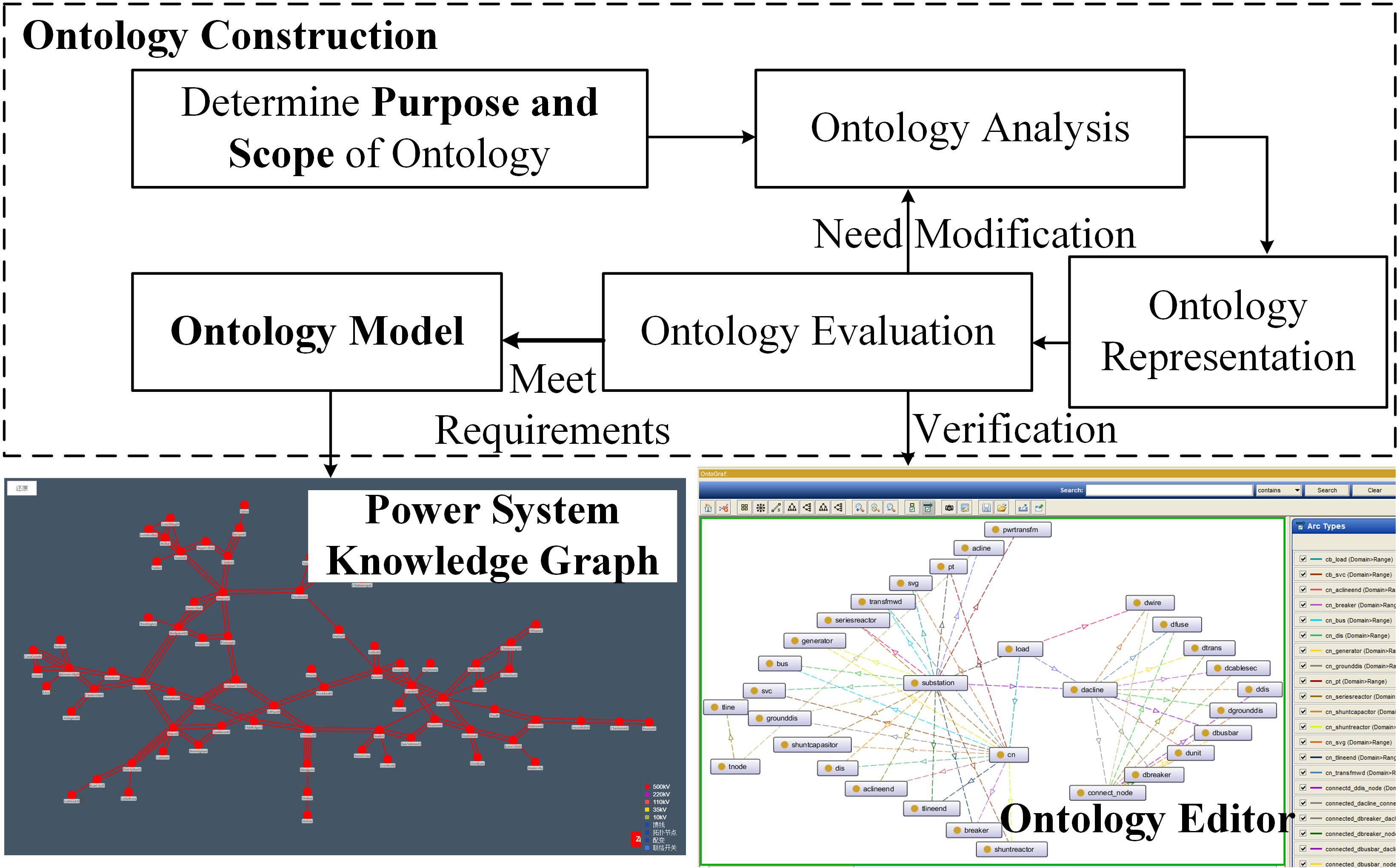}}
\caption{Example of Ontology Modeling and Application for Power System.}\label{figure2}
\end{figure}

Fig. \ref{figure2} demonstrates the example of ontology modeling and application for the power system. The ontology construction proposed in this paper adopts a top-down approach, which follows the principles of clarity, coherence, extendibility, minimal encoding bias, and minimal ontological commitment. As shown in this figure, the authors use the ontology editor \cite{17} to verify the constructed ontology model. Building an ontology model for the power system based on a graph database can fully use the natural consistency between the physical topology in the power grid and the ``node'' and ``edge'' data structure in the graph database. The constructed ontology model can form the power knowledge graph more intuitively through entity loading. The ontology model should include full-service management modules such as power grid topology, power equipment management, customer service, power market, and power generation.

\section{Construction of IQA System for Electric Power Knowledge Graph}
\subsection{Semantic Structure Analysis of Questions}
The analysis accuracy of questions that representing users' intention will directly determine the answers' precision from the IQA system. In terms of domain knowledge, questions will involve strong professionalism, more complex constraints, and higher requirements for accuracy. The proposed method in this paper analyzes the semantic structure of the problem in-depth, determines the constraints and question goals given by users, and realizes the highly accurate analysis of the users' intention. The specific steps are as follows.

\subsubsection{Entity Recognition}
The IQA system first uses NLP to segment words for question sentences and performs entity recognition based on word segmentation, which will form an entity tag set $E$. The entities in a question usually include both electric power professional words and commonly used words. The ontology model and entities can effectively cover the related electric power professional words in the established electric power knowledge graph. For general vocabularies, this method integrates the thesaurus of systems such as iFLYTEK and Jieba to cover the common vocabulary. The part-of-speech tagging in the question are performed through word segmentation, and the extracted words (entities) are labeled simultaneously. The label is the name of the node, edge, or attribute of the ontology model. This algorithm then ignores connectives and non-entity words and automatically through the ontology model's logical association relationship to determine the problem type and keywords. Among them, for the tags of the same entity, a specific vocabulary is supplemented, including multiple expressions of one statement, such as an organization described by the full name or the corresponding abbreviation.

\subsubsection{Target Extraction}
\begin{figure}[h!bt]
\centerline{\includegraphics[scale = 0.25] {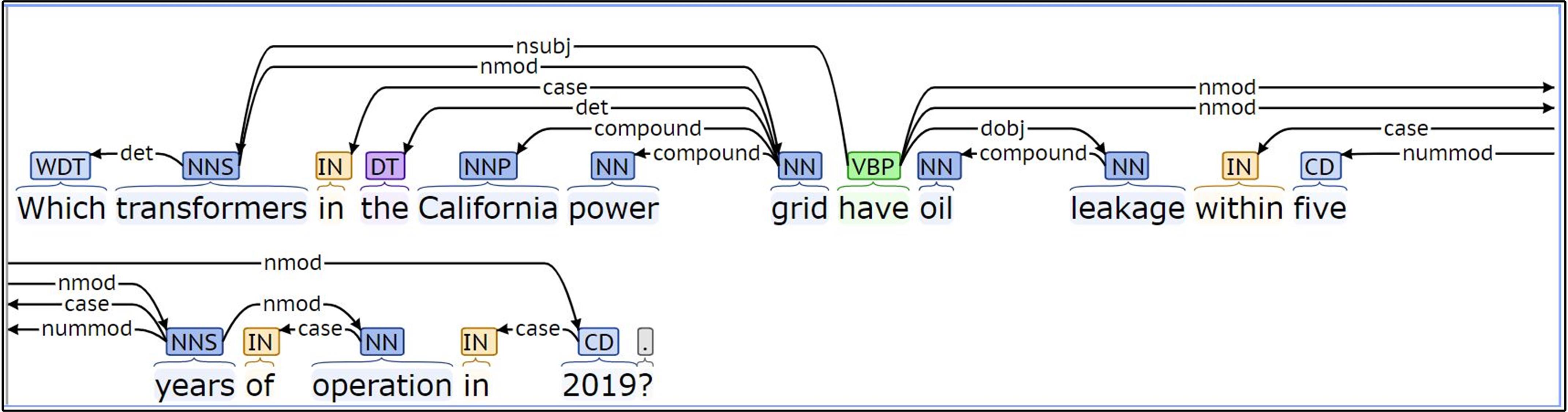}}
\caption{Example of Dependencies Tree.}\label{figure3}
\end{figure}

According to the entity-tag set $E$, the IQA system uses dependency tree analysis to extract the question target $T$. The dependency tree is a method of representing the structure of words or phrases in a sentence according to the grammatical or semantic relationship. As shown in Fig. \ref{figure3}, the question "Which transformers in the California power grid have oil leakage within five years of operation in 2019?" is a dependency tree formed after semantic structure analysis using Stanford University's open-source software CoreNLP \cite{18}. In the basic dependencies, the ``California'' is recognized as the state or province and ``power grid'' is compound with ``California'', ``five years'' is a time duration, and ``2019'' is the date. Besides, ``transformers'' is identified as the nominal subject (nsubj), and some other relationships such as direct object (obj) and conditions (case) are also illustrated in the tree. Here, NN represents noun, DT is determiner, VE is verb, and PN means proper noun.

\begin{figure}[h!bt]
\centerline{\includegraphics[scale = 0.33] {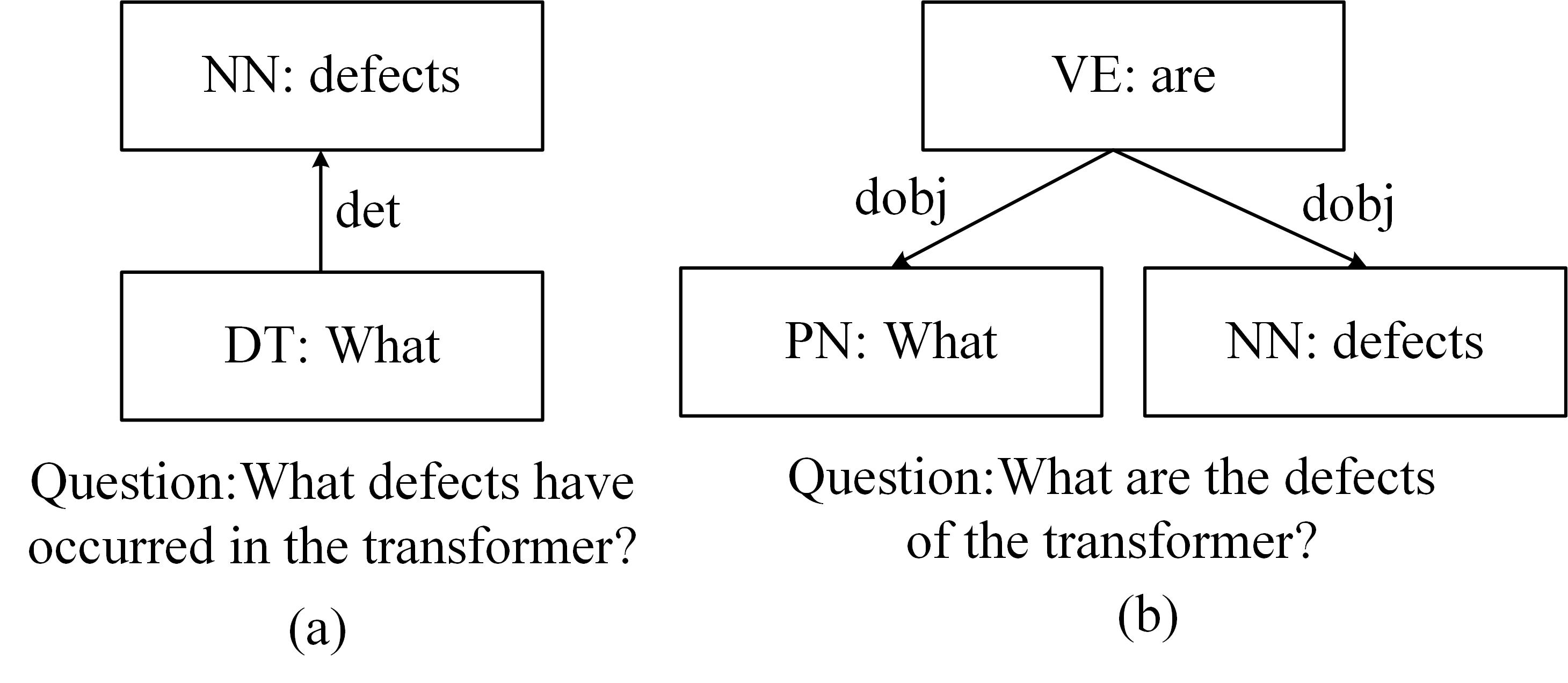}}
\caption{Dependencies Tree Structure of Question Target.}\label{figure4}
\end{figure}

By analyzing the dependency tree, it can be found that the question target related to the power system can usually be parsed into two dependency tree structures, as shown in Fig. \ref{figure4}: (a) Parent-child structure and (b) Brothers structure. In the parent-child format, the parent node is the target node, and the child node is the problem type. The brothers structure often appears in inverted questions. The common parent node is a verb, and children nodes have question types and target nodes, and some items also include time, place, and other qualifiers for the action.

\subsubsection{Constraints Extraction}
According to the entity-tag set $E$, the constraint set $C$ can be extracted from the question. The relationship between time and logical words in common vocabulary and other entity-tags can be determined through the dependency tree. For example, ``operation time'' and ``within'' respectively indicate the relationship between time qualifiers and logical qualifiers with the subject (transformer).

\subsection{Reasoning Algorithm based on Ontology Model}
The knowledge reasoning algorithm designed in this paper for IQA is a graph shortest path method that integrates multiple rules. It determines the shortest path between all constraints nodes and the target node. The shortest path satisfies the laws in all constraints, such as the selection of multiple edges between nodes and/or logical combination judgment (which include ``and'', ``not'', ``or'').

\begin{algorithm}[h!tb]
\caption{Reasoning Algorithm for IQA System}
\label{algorithm1}
\begin{algorithmic}
\REQUIRE ~~\\
$E$: Ontology Set.\\
$T$: Question Target.\\
$C$: Constraint Set, $C$ = \{$C_1, C_2, \cdots, C_i, \cdots, C_n$\}.\\
\renewcommand{\algorithmicrequire}{ \textbf{Input:}}
\REQUIRE {$T$, $C$}\\  
\renewcommand{\algorithmicensure}{ \textbf{Iteration Process:}}
\ENSURE {Select $C_i$ from $C$ to match $E$.}\\
\STATE Perform DFS/BFS/Dijkstra to find the shortest path $P$.
\IF {\{$C_1, C_2, \cdots, C_{i-1}, C_{i+1}, \cdots, C_n$\} on $P$}
\RETURN $P$;
   \ELSE
   \STATE $C$ $\leftarrow$ \{$C_1, C_2, \cdots, C_{i-1}, C_{i+1}, \cdots, C_n$\};
   \STATE go to \textbf{Iteration Process}
\ENDIF
     \end{algorithmic}
\end{algorithm}

\begin{figure}[h!bt]
\centerline{\includegraphics[scale = 0.24] {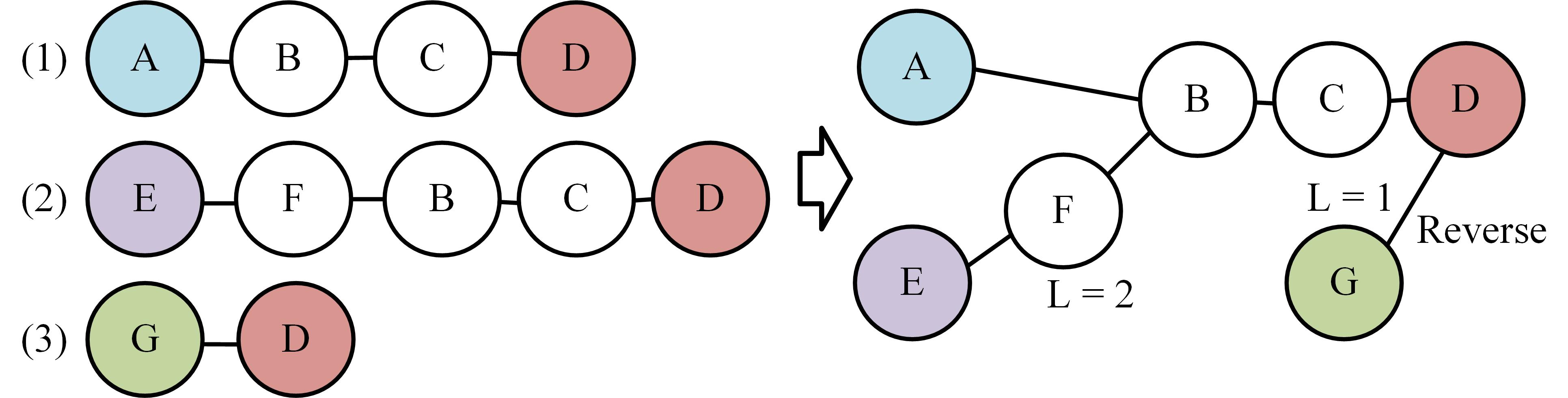}}
\caption{Intersection of Shortest Paths.}\label{figure6}
\end{figure}

The pseudocode of the proposed reasoning algorithm is shown as Alg.(\ref{algorithm1}) and the specific descriptions and announcements are as follows:
\begin{enumerate}
    \item Randomly select the first constraint and execute the shortest path algorithm. The algorithm library contains DFS, BFS, Dijkstra, etc., to find the corresponding most straightforward path;
    \item Search for other constraints. If there are related nodes or edges in the existing shortest path, add the condition directly to the current shortest path; Otherwise, perform the shortest path search again and continuously judge whether the current shortest path already exists during the search process: If it exists, the search is interrupted, and the existing most straightforward approach is introduced as the shortest path of the constraint; If not, the search continues until determining the target node;
    \item Determine whether there are two or more shortest paths in all the search results that meet the constraints, and at the same time, judge whether there are other common nodes in the path except the target node. If they exist, intersect the shortest path of these nodes and target node. As shown in Fig. \ref{figure6}, there are three shortest paths for three different constraints (A, E, G). The three paths are intersected as one route contains all restrictions.  Regarding the $L = 1$ (G to D) intersection, according to the graph database characteristics, set a reverse edge to this path instead of adding a new vertex, which can accelerate query generation and search efficiency. 
\end{enumerate}

Therefore, the query statement of the graph database can be formed according to the shortest path. Implementing the ontology model to organize a large number of scattered knowledge in an orderly manner: the highly relevant ontology is designed as the core node; other ontologies are aggregated on the core node as attributes or edge attributes, which simplifies the complexity of the knowledge graph and significantly reduces graph traversal hop count of the algorithm during reasoning.

\subsection{System Design}
According to the above method, we have implemented a complete set of IQA system based on a power knowledge graph. The back end of the system is responsible for question analysis and knowledge reasoning, and the front end is responsible for user interaction and knowledge display. For user convenience, we set a standard format for questions in our user manual. The parsing results from different expressions of the same question might be slightly different, but the system can return the proper answer if they can match the ontology. It can be used as a plug-in and integrated into any front-end pages. When users browse the main page, they often want to further search and reveal relevant knowledge. The IQA system will help them obtain relevant knowledge dynamically in real-time. 

\begin{figure}[h!bt]
\centerline{\includegraphics[scale = 0.6] {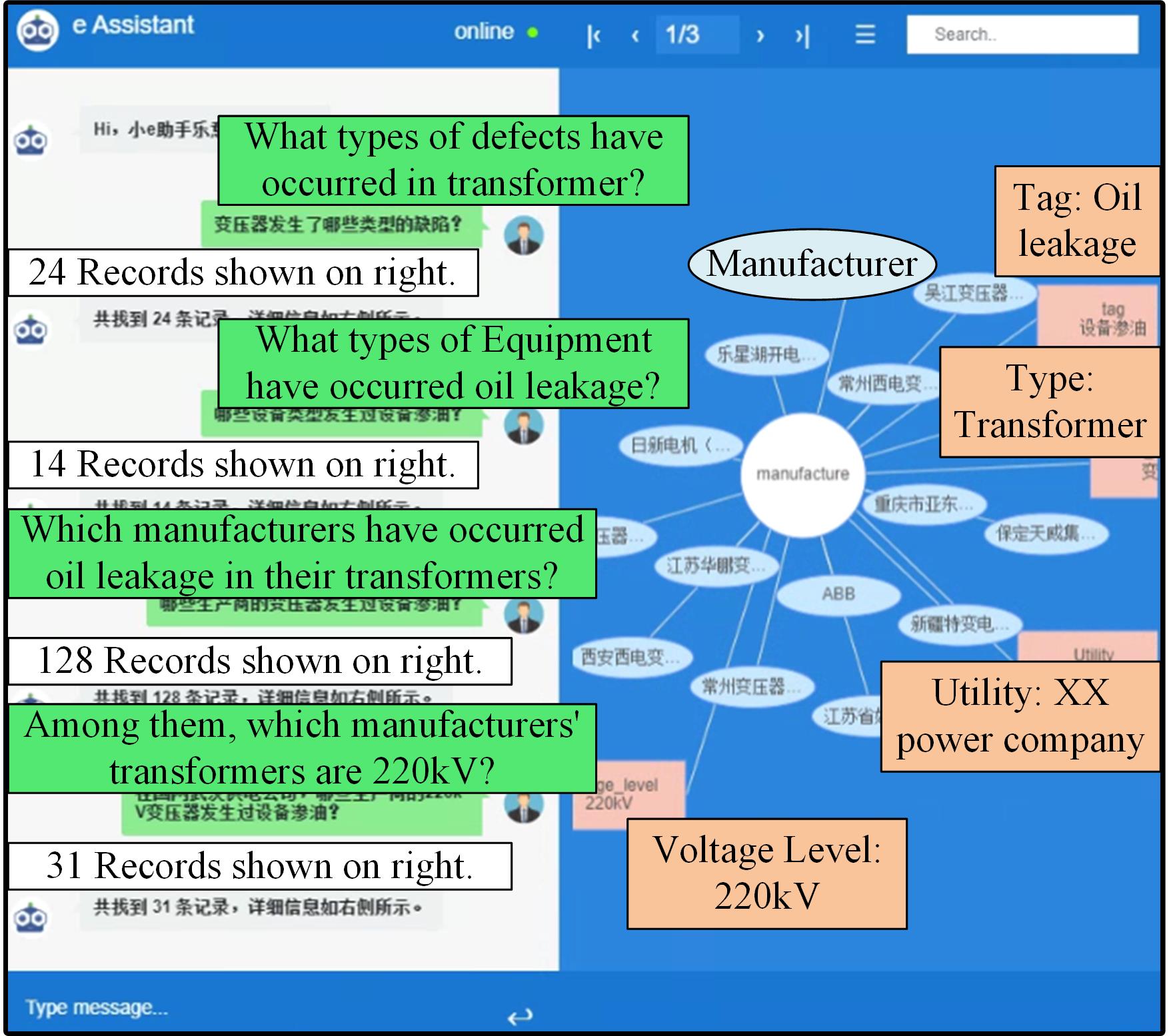}}
\caption{Demonstration of IQA system.}\label{figure5}
\end{figure}

As shown in Fig. \ref{figure5}, the goals and constraints corresponding to the question, as well as all the relevant knowledge acquired in the knowledge base, can be visually displayed in the graph on the right. As illustrated in this example, the last question is the extension of the previous question. The current question's target node is still the manufacturer, and the constraints that include the oil leakage tag and the transformer equipment type are kept. Another filter condition: voltage level as 220kV is added from the new question. All the related restrictions and returned results are displayed on the right side of the IQA system in a knowledge graph form. Simultaneously, other constraints related to the search results, such as utility, are also displayed in the graph to facilitate further searches and questions. The user can also select any knowledge node as the starting condition, from which to start new questions; The IQA system can use the same algorithm to obtain answers.

\section{Case Study}
To verify the accuracy and comprehensiveness of the IQA system regarding identifying questions and reply answers, we built an experimental environment and tested the system. The testing environment uses a computer server with an 8-core CPU and 16G memory. The graph database uses Tigergraph Database version 3.0. The constructed power knowledge graph's ontology model contains 15 ontologies, including physical equipment, voltage level, manufacturer, equipment type, utilities, etc. The data scale reaches more than 4 million nodes and more than 38 million edges. Since there is no public test set and evaluation method for question and answering system in the electric power field, based on the existing knowledge graph, we use the expert system to test 200 common problems in the power grid management field generated by the questionnaire. These questions relate to 10 ontologies and their corresponding relationships. In terms of graph traversal, the tested queries cover 10 single-hop and 190 multi-hop queries; in terms of restriction conditions, the tested queries cover 20 single-condition and 180 multi-condition queries. 

Table \ref{case} shows the test results from question analysis, reasoning, feedback answer, and execution time.

\begin{table}[h!t]
\caption{Test Results.}
\label{case}
\newcommand{\tabincell}[2]{\begin{tabular}{@{}#1@{}}#2\end{tabular}}
\centering 
\begin{tabular}{c | c | c | c | c | c } 
\hline\hline
\textbf{Category} & \textbf{\tabincell{c}{Single\\-hop}} & \textbf{\tabincell{c}{Multi\\-hop}} & \textbf{\tabincell{c}{Single\\-condition}} & \textbf{\tabincell{c}{Multi\\-condition}} & \textbf{Total}\\ \hline
\textbf{Quantity} & 10 & 190 & 20 & 180 & 200 \\ \hline
\textbf{\tabincell{c}{Parsing\\ Error}} & 0 & 3 & 1 & 2 & 3 \\ \hline
\textbf{\tabincell{c}{Reasoning\\ Error}} & 0 & 2 & 0 & 2 & 2 \\ \hline
\textbf{\tabincell{c}{Accepted\\ Answers}} & 100\% & 98.42\% & 95.00\% & 98.86\% & 98.50\% \\ \hline
\textbf{\tabincell{c}{Average\\ Execution \\Time}} & 162ms & 614ms & 153ms & 637ms & 588ms\\ \hline\hline
\end{tabular}
\end{table}

According to the above results, the IQA system works well in the experimental environment. The question answers are compared with the expert system. The average question resolution accuracy rate is close to 99\%, the average answer accuracy rate is close to 98\%, and the average response time is less than 600ms. The parsing errors are mainly due to three words in questions are not involved in the initial custom dictionary. Therefore, it can be verified that the system can accurately answer users' natural questions in real-time, and the algorithm is feasible. 

We further analyzed the wrong answers. There were three question parsing errors, which directly led to the wrong final answers. Besides, the reasoning error was also because the question target can not be parsed. Therefore, by iteratively improving the knowledge base and entity identification mechanism, the accuracy of the IQA system can be continuously improved.

\section{Conclusion and Future Work}
This paper introduces an intelligent question answering system based on the power knowledge graph. The design makes full use of ontology models and natural language processing technology to quickly and accurately extract the constraints and goals of questions. The graph database query is then automatically formed through the designed reasoning algorithm, and the knowledge of the corresponding answer can be searched and obtained from the knowledge graph. Because the ontology in the domain knowledge graph is usually highly professional and cohesive, it is easy to optimize and form an ontology model with low complexity. Therefore, using the optimized ontology model, the number of hops in a single search can be controlled within the graph's diameter, which can reduce the complexity of the reasoning algorithm and can also efficiently query and analyze the immense knowledge in the power field. 

We found that using dependency trees to analyze objectives and constraints has high accuracy in expressing logically straightforward questions. However, the randomness of the user's query will reduce accuracy of the answer, such as the omission of logical words. Meanwhile, some business knowledge needs to be obtained by reasoning through multiple rounds of questions with complex logic levels. In future work, we plan to combine the context to design the detection method for further question intention based on different business scenarios and provide question recommendations. To improve question parsing accuracy, we will offer users question templates to standardize the query format. For questions that cannot be resolved and answered, the system plans to design a question return mechanism to facilitate the system's follow-up improvement and the filling of the knowledge base.

\bibliographystyle{IEEEtran}
\bibliography{IEEEabrv,RefDatabase}

\end{document}